Chinese Intermediate English Learners outdid ChatGPT in deep cohesion: Evidence from English narrative writing

**Tongquan Zhou[1*], Siyi Cao[1*/**], Siruo Zhou[2**], Yao Zhang[1], Aijing He[3**]**

1. School of Foreign Languages, Southeast Univesity, Nanjing, China, 211189

2. School of Foreign Studies, Nanjing University of Posts and Telecommunications, Nanjing, China 210023

3. School of Foreign Studies, Guangzhou University, Guangzhou, China 510006

\* Equal contribution, sharing first authorship

**\*\*Corresponding authors:** siyi.c@nuaa.edu.cn, susanzhou@naver.com, sfsheaijing@gzhu.edu.cn

**Abstract**

ChatGPT is a publicly available chatbot that can quickly generate texts on given topics, but it is unknown whether the chatbot is really superior to human writers in all aspects of writing and whether its writing quality can be prominently improved on the basis of updating commands. Consequently, this study compared the writing performance on a narrative topic by ChatGPT and Chinese intermediate English (CIE) learners so as to reveal the chatbot's advantage and disadvantage in writing.

The data were analyzed in terms of five discourse components using Coh-Metrix (a special instrument for analyzing language discourses), and the results revealed that ChatGPT performed better than human writers in narrativity, word concreteness, and referential cohesion, but worse in syntactic simplicity and deep cohesion in its initial version. After more revision commands were updated, while the resulting version was facilitated in syntactic simplicity, yet it still lagged far behind CIE learners' writing in deep cohesion. In addition, the correlation analysis of the discourse components suggests that narrativity was correlated with referential cohesion in both ChatGPT and human writers, but the correlations varied within each group.

## 1.Introduction

ChatGPT, created by OpenAI, is an advanced chatbot that can fulfill a wide range of text-based requests. It can answer simple questions, generate complex documents like "thank you" letters, and guide individuals through discussions such as programming code. The chatbot uses its extensive data stores and efficient design to comprehend user requests and generate responses in almost natural languages (De et al., 2023; Liu et al., 2021).

ChatGPT is built on GPT3.5, which is one of the largest language models (LLMs). LLMs are deep learning models that can understand and create natural language, and they undergo a two-step refinement process. The first step involves generative, unsupervised pretraining that uses unlabeled data to refine the model's comprehension of language. The second step is discriminative, supervised fine-tuning that enhances the model's performance on specific tasks. This approach allows LLMs to become more specialized and optimized for particular uses (Brown et al., 2020; Lund & Wang, 2023; Min et al., 2021; Radford et al., 2018; Vaswani et al., 2017).

ChatGPT was trained on an extensive collection of diverse texts from the internet, including books, articles, and websites, covering a wide range of topics such as news, Wikipedia, and fiction. Additionally, ChatGPT was specifically fine-tuned for conversational applications using reinforcement learning with human feedback. This method enables ChatGPT to adjust its responses based on human feedback, enhancing its ability to understand user intentions, create human-like text, and maintain coherence in a conversation. As a result, ChatGPT is highly proficient in carrying out conversational tasks (Shen et al., 2023).

ChatGPT's real-world applications are impressive, but the chatbot has sparked numerous linguistic debates. Many well-known linguists oppose ChatGPT's approach. For instance, Noam Chomsky posits that each individual possesses a *linguistic competence*, an innate ability to comprehend and produce grammatically correct sentences in their language. This means that a person's understanding of their language's grammar and structures, as well as their ability to apply this knowledge to understand and produce sentences, is not acquired through imitation or reinforcement but rather an innate ability that is hardwired into the human brain (Abdulrahman et al., 2019). Similarly, Steven Pinker has suggested the idea of *language instinct*, claiming that humans have an innate capacity for language acquisition that is genetically determined and part of which makes humans unique as a species (Pinker, 2003). According to these theoretical views, ChatGPT lacks the innate language abilities to acquire and use language but rather emulates the patterns it "sees" in data.

However, two recent studies have unveiled remarkable language capabilities of ChatGPT. In one study, Benzon (2023) created a novel approach based on Chomsky's linguistic competence and David Marr's three levels of analysis for information systems (Marr, 1982) to examine ChatGPT's abilities. The results indicated that ChatGPT had a higher command of sophisticated discourse skills, among which are performing analogical reasoning tasks, interpreting films and stories, and even comprehending abstract concepts like justice and charity. In another study, Kosinski (2023) found that ChatGPT was comparable to a nine-year-old child in the performance on theory of mind (ToM) tasks. For example, when presented with a stimulus saying "John put the cat in the basket and left. Mark took the cat out of the basket and put it in the box while John was away," ChatGPT was able to determine the "cat's location" and "where John would look for the cat when he returned".

Remarkably, ChatGPT succeeded in completing all 20 similar tasks with 100% accuracy.

Despite these impressive findings, some experts in the field of artificial intelligence (AI) have cast doubt on the validity of these experimental results. For instance, Songchun Zhu, a professor at Peking University (China) argued that ChatGPT's ability to pass ToM tasks merely demonstrated its capacity to pass a test of ToM and did not necessarily imply that it has ToM. Moreover, he raised some challenging questions about the validity of traditional testing tasks for machines' ToM development and their ability to complete these tasks without having ToM. Another professor from UC Beckley, Stuart Russell, also questioned the existence of ChatGPT's ToM, claiming that ChatGPT lacked the ability to learn and express complex generalizations, as evidenced by the need for a large amount of textual data and still producing errors (Synced, 2023).

English writing is considered an important skill in our daily lives (Petchprasert, 2021). In the past, the writing has been made by human writers, and this situation changed until AI writers like ChatGPT came to our lifes. As repetitively mentioned in literature, ChatGPT has the potential to provide many benefits in this area, including language assistance, translation, editing, and proofreading (e.g., Jiao et al., 2023; M Alshater, 2022). For example, ChatGPT can enhance researchers' language proficiency and writing skills by providing real-time feedback on grammar, syntax, spelling, and vocabulary, thereby increasing the overall quality of their manuscripts (Bishop, 2023). Additionally, ChatGPT has the capability to promptly generate text on complex or technical topics (Lund & Wang, 2023), although Chomsky has deemed that ChatGPT is essentially high-tech plagiarism. One of the key features of ChatGPT is its personalized prompt strategy, which generates prompts tailored to the user's interests and needs (White et al., 2023). This allows ChatGPT to revise the manuscript according to the user's needs. To date, nevertheless, no research has been conducted to explore the quality of English writing produced by ChatGPT or whether its writing quality can be improved so as to correspond to user responses.

In the context of English as a foreign language learning, English writing is generally considered an important skill for English majors. According to Jin and Fan (2011), Chinese English majors who pass the Test for English Majors-Band 4 (TEM4) are considered to have intermediate proficiency in English. In fact, research has shown that Chinese English majors are able to produce grammatically complex English argumentative writing that is similar to EFL textbooks (Shao et al., 2022). Additionally, Chinese English majors tend to make few errors in their English writing (e.g., 5.99% in Chinglish), with spelling errors being the most common error type (29.3%), often due to carelessness (Sun & Shang, 2010). Against this background, an interesting issue is whether Chinese intermediate English majors as human writers do better than ChatGPT in terms of English writing proficiency.

Narrative writing involves the creative and lifelike portrayal of events, experiences, or emotions through the use of characters, plot, setting, and dialogue. It serves as a means of personal expression and entertainment (Bold, 2011). During a narrative writing, authors demonstrate their ability to narrate by exploring the motivations and perspectives of characters. Moreover, it necessitates critical thinking and logical reasoning, as writers must carefully consider elements like plot, pacing, and character development (Holloway et al., 2007). Finally, narrative writing requires language proficiency and the use of descriptive and engaging language (Ellis & Yuan, 2004). Given that narrative writing demands a greater deal of critical thinking, logical reasoning, and language skills in the second language, it is rational to use the writing as an window to measure and assess the writing quality of CIE learners.

    Natural Language Processing (NLP) vehicles are widely used to evaluate and analyze language discourses in texts. Coh-Metrix is one such tool that has been extensively employed in L2 writing research to measure discourse components within the text (McNamara et al., 2014). For instance, Varner et al. (2013) used Coh-Metrix as the criteria by teachers and students to evaluate their discrepancies in writing quality. Graesser et al. (2014) reviewed how Coh-Metrix accounted for text variations in terms of five discourse components (i.e., narrativity, syntactic simplicity, word concreteness, referential cohesion, and deep cohesion). Similarly, Petchprasert (2021) found that the factors of word concreteness, referential cohesion, and deep cohesion had an impact on students' writing performance. In this study, Coh-Metrix was adopted to compare the writing performance of ChatGPT and CIE learners based on five discourse components: narrativity, syntactic simplicity, word concreteness, referential cohesion, and deep cohesion, as described in Table 1.

Table 1. The description of five discourse components (Petchprasert,2021)

| **Discourse components** | **Description** |
|---|---|
| Narrativity | Stories typically involve characters, events, places, and objects that are recognizable and relatable to readers, conveying a narrative that is reminiscent of everyday spoken discourse. |
| Syntactic simplicity | This component measures the extent to which sentences in the text feature simple, commonly used syntax or more complex, unfamiliar syntax with longer or shorter word length. |
| Word concreteness | Content words are tangible, significant, and straightforward to comprehend, whereas abstract words are more challenging to represent visually, making texts with a higher frequency of abstract words more difficult to understand than those with content words. |
| Referential cohesion | Greater referential cohesion in a text means that the words and concepts used extend beyond individual sentences and the overall text, resulting in a more coherent and connected piece. In contrast, lower cohesion tends to make it more challenging to comprehend and link together the ideas presented in the text. |
| Deep cohesion | This component indicates the extent to which the text includes causal and intentional connectors that aid readers in developing a more comprehensive and interconnected understanding of the events, processes, or actions described in the text. |

To summarize, previous research has mainly achieved four accomplishments. Firstly, ChatGPT has been proven to be powerful in English writing. Secondly, the chatbot can generate numerous ideas based on given topics. Thirdly, ChatGPT has a prompt strategy that generates new prompts based on user responses. Fourthly, CIE learners as human writers have shown good writing performances, particularly in narrative writing.

Despite these accomplishments, it remains unknown whether ChatGPT performs better than CIE learners in English narrative writing. Therefore, the present study aims to use Coh-Metrix to answer the following questions regarding narrative writing:

(1) How do ChatGPT and CIE learners perform in narrative English writing from the perspective of discourse components?

(2) Which of the components in a revised version can be improved by ChatGPT when user responses are updated for revision?

(3) How are the discourse components of the texts produced by ChatGPT and CIE learners correlated?

**2. Method**

2.1 Participants

Forty sophomores majoring in English (10 males and 30 females) (native Chinese speakers) in a top-20 university in China voluntarily participated in our study and their age ranged from 18 to 20 ($M = 18.35$, $SD = 1.67$). Although they had not taken TEM4 examination (the next semester) prior to the experimental writing, they were evaluated by the university in terms of English listening, speaking, reading and writing before enrollment. More importantly, they have learned English more than one year after enrollment, which indicates that they have already demonstrated an intermediate level of English proficiency.

In this study, we adopted the paradigm described by Petchprasert (2021) and the experiment was approved by the Human Research Ethics Committee of the university with which the first author is affiliated.

2.2 Procedure

As a part of this study, all 40 participants, consisting of 10 males and 30 females, were given instructions to write a narrative essay on the topic "*Write a personal experience*" within 500 words. The participants were required to submit their essays in digital form within a week. To ensure fairness, the participants were informed that their final scores would not be influenced by their performance on this task, and hence they were prohibited from using automatic grammar and spelling checker tools. For the sake of comparison, ChatGPT was used to generate 40 English writings on the same topic, ensuring that no content was duplicated between them.

After comparing the initial writing from ChatGPT and CIE learners, we gave a command to ChatGPT to "*revise the manuscript in terms of detailed discourse components (e.g., deep cohesion)*" based on the results of comparison.

2.3 Coding and data analysis

Coh-Metrix is a comprehensive text analysis tool and one of its unique features is the ability to analyze both surface-level and deep-level linguistic features of a text. For example, Coh-Metrix can identify and quantify rhetorical and discourse features such as narrative structure, and coherence, which are important for understanding how a text is organized and how it communicates its message (McNamara et al., 2014). Therefore, all collected English writings were first analyzed by Coh-Metrix 3.0 in terms of the five discourse components (i.e., narrativity, syntactic simplicity, word concreteness, referential cohesion, and deep cohesion). Then, the statistical data was analyzed by multivariate analysis of variance (MANOVA) and correlation using R (R Core Team, 2016). Specifically, MANOVA was applied by *manova* function in R. When a significant main effect was observed, the follow-up tests of one-way ANOVAs were conducted by using EMMEANS function in bruceR package and then Tukey's method for multiple comparisons was utilized. Correlation was adopted by PerformanceAnalytics package.

3. Results

3.1 Comparison between ChatGPT's first version and CIE learners' wrting

Table 2 demonstrates the percentile (i.e., mean and SD) of ChatGPT and CIE learners in the five discourse components, including narrativity, syntactic simplicity, word concreteness, referential cohesion, and deep cohesion. A multivariate analysis of variance (MANOVA) was conducted to test whether ChatGPT did better than CIE learners in the first version of narrative writing. All aforementioned variables were included as dependent variables with the factor of Group ( ChatGPT and CIE learners ) as independent variables. The results showed an overall significant main effect of "Group" ($F(5, 74) = 11.545$, $p < .001$). Further, one-way ANOVA on each dependent variable was conducted as follow-up tests to the MANOVA, revealing the main effect of "Group" in terms of narrativity ($F(1, 78) = 4.608$, $p < .05$), syntactic simplicity ($F(1, 78) = 6.262$, $p < .05$), word concreteness ($F(1, 78) = 12.380$, $p < .001$), referential cohesion ($F(1, 78) = 434.327$, $p < .001$), and deep cohesion ($F(1, 78) = 11.868$, $p < .001$). As displayed in Figure 1, the multiple comparisons using Tukey method showed that ChatGPT had higher percentile than CIE learners in narrativity (*β*(Chinese - ChatGPT) = -3.683, *t*(78) = -2.147, *p* < .05), word concreteness (*β*(Chinese - ChatGPT) = -17.834, *t*(78) = -3.519, *p* < .001) and referential cohesion (*β*(Chinese - ChatGPT) = -25.724, *t*(78) = -5.859, *p* < .001). Conversely, CIE learners had higher percentile than ChatGPT in syntactic simplicity (*β*(Chinese - ChatGPT) = 9.278, *t*(78) = 2.502, *p* < .05) and deep cohesion (*β*(Chinese - ChatGPT) = 16.308, *t*(78) = 3.445, *p* < .001). This indicates that in the first version, ChatGPT performed better than CIE learners in narrativity, word concreteness and referential cohesion, but worse in syntactic simplicity and deep cohesion.

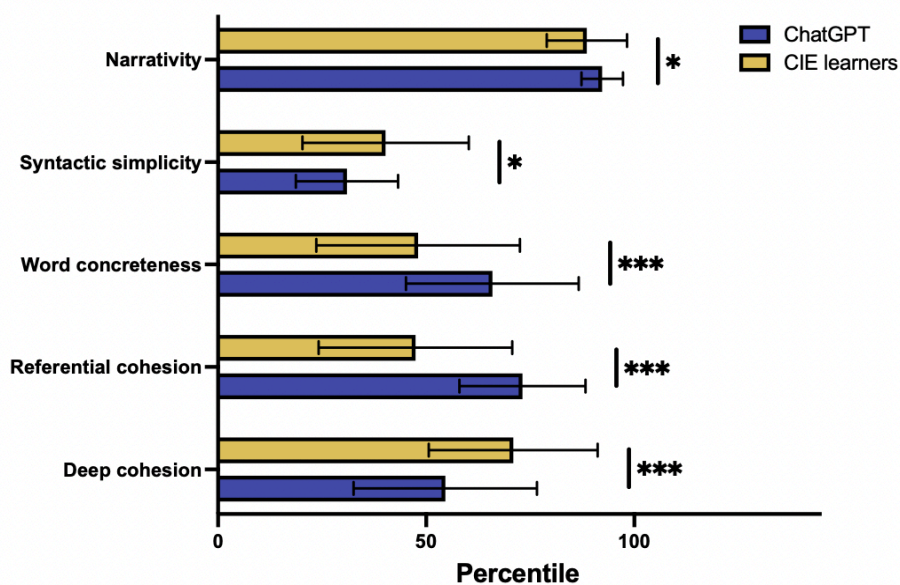

Figure 1. Percentile of five components of English writing by ChatGPT and CIE learners

Table 2. Five components of the first version by ChatGPT and CIE learners

| Variable | ChatGPT | | CIE learners | | *F*(1, 78) | *p* | $\eta^2 p$ |
| --- | --- | --- | --- | --- | --- | --- | --- |
| | *M* | *SD* | *M* | *SD* | | | |
| Narrativity | 92.294 | 4.983 | 88.611 | 9.640 | 4.608 | .035 | .056 |
| Syntactic simplicity | 30.974 | 12.270 | 40.252 | 19.982 | 6.262 | .01 | .074 |
| Word concreteness | 65.904 | 20.749 | 48.070 | 24.435 | 12.380 | <.001 | .137 |
| Referential cohesion | 73.155 | 15.150 | 47.431 | 23.271 | 34.327 | <.001 | .306 |
| Deep cohesion | 54.610 | 22.030 | 70.918 | 20.276 | 11.868 | <.001 | .132 |

## 3.2 Comparison between ChatGPT's writing (two versions) and CIE learners' writing

Table 3 and Figure 2 display the average and standard deviation of five discourse components (narrativity, syntactic simplicity, word concreteness, referential cohesion, and deep cohesion) in the first and the revised version of writing by ChatGPT and the writing by CIE learners. A multivariate analysis of variance (MANOVA) was conducted to compare their performance. The dependent variables included all five discourse components, and the independent variable was the group (i.e., the first version by ChatGPT, the revised version by ChatGPT and CIE learners' writing). The results indicated a significant main effect of the group factor ($F(228, 10) = 5.30$, $p < .001$). Further analysis using one-way ANOVA revealed the main effect of "Group" in terms of narrativity ($F(2, 117) = 3.539$, $p < .05$), syntactic simplicity ($F(2, 117) = 3.654$, $p < .05$), word concreteness ($F(2, 117) = 8.357$, $p < .001$), referential cohesion ($F(2, 117) = 17.372$, $p < .001$), and deep cohesion ($F(2, 117) = 5.878$, $p < .01$). Specifically, in narrativity component, ChatGPT exhibited similar performance in both the first and revised version of writing, compared with CIE learners ($β$(Chinese – ChatGPT1) = -3.683, $t(117) = .-1.983$, $p = .149$; $β$(Chinese – ChatGPT2) = 1.012, $t(117) = .545$, $p = 1.000$), although ChatGPT performed poorer in the revised version than in the first version ($β$(ChatGPT2 – ChatGPT1) = -4.695, $t(117) = -2.527$, $p = .038$).

In syntactic simplicity, CIE learners outdid ChatGPT in the first version ($β$(Chinese – ChatGPT1) = 9.278, $t(117) = 2.644$, $p = .028$), but showed no difference from ChatGPT in the revised version ($β$(Chinese - ChatGPT2) = 6.353, $t(117) = 1.810$, $p = .218$).

In word concreteness, although there appeared no improvement between the first and revised version of writing by ChatGPT ($β$(ChatGPT2 – ChatGPT1) = .904, $t(117) = .175$, $p = 1.000$), ChatGPT in the two versions exceeded CIE learners ($β$(Chinese – ChatGPT1) = -17.834, $t(117) = -3.450$, $p = .002$; $β$(Chinese – ChatGPT2) = -18.737, $t(117) = -3.625$, $p = .001$).

In referential cohesion, ChatGPT showed poorer performance in the revised version than in the first version ($β$(ChatGPT2 – ChatGPT1) = -11.011, $t(117) = -2.514$, $p = .040$), but displayed better performance than CIE learners ($β$(Chinese – ChatGPT1) = -25.724, $t(117) = -5.874$, $p < .001$; $β$(Chinese – ChatGPT2) = -14.713, $t(117) = -3.360$, $p = .003$). Conversely, the unexpected results in deep cohesion showed that CIE learners surpassed ChatGPT in the two versions ($β$(Chinese – ChatGPT1) = 16.308, $t(117) = 3.304$, $p = .004$; $β$(Chinese – ChatGPT2) = 12.068, $t(117) = 2.445$, $p = .048$; $β$(ChatGPT2 – ChatGPT1) = 4.240, $t(117) = .859$, $p = 1.000$).

In summary, ChatGPT in the revised version kept doing better in word concreteness and referential cohesion than CIE learners, but worse in deep cohesion.

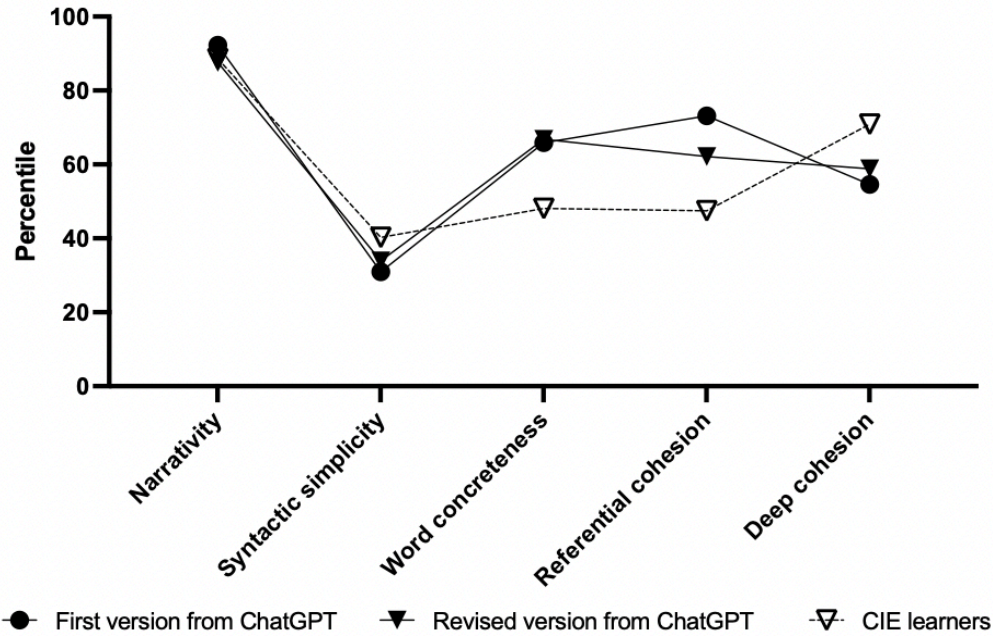

Figure 2. Percentile of five components of English writing ChatGPT's writing (two versions) and CIE learners' writing

## 3.4 Correlations of five components between ChatGPT's revised version and CIE learners' writing

In order to examine whether there is any significant correlation between those discourse components across ChatGPT and CIE learners, the correlation analysis was conducted. As illustrated in Table 4 and Figure 3, for ChatGPT group, narrativity was positively correlated with referential cohesion ($r = .64$, $p < .001$), and word concreteness was negatively correlated with syntactic simplicity ($r = -.38$, $p < .05$), but positively correlated with referential cohesion ($r = .38$, $p < .05$). For CIE learners, narrativity was also significantly correlated with referential cohesion ($r = .49$, $p < .01$) and syntactic simplicity was negatively correlated with referential cohesion ($r = -.45$, $p < .01$).

Table 4. Correlations separately for ChatGPT (above) and CIE learners (below)

| Variable | 1 | 2 | 3 | 4 | 5 |
|---|---|---|---|---|---|
| 1. Narrativity | - | -.01 | -.06 | .64*** | .12 |
| 2. Syntactic simplicity | -.09 | - | -.38* | -.25 | .17 |
| 3. Word concreteness | -.22 | -.26 | - | .38* | .074 |
| 4. Referential cohesion | .49** | -.45** | -.11 | - | .06 |
| 5. Deep cohesion | .28 | .0058 | .0033 | .036 | - |

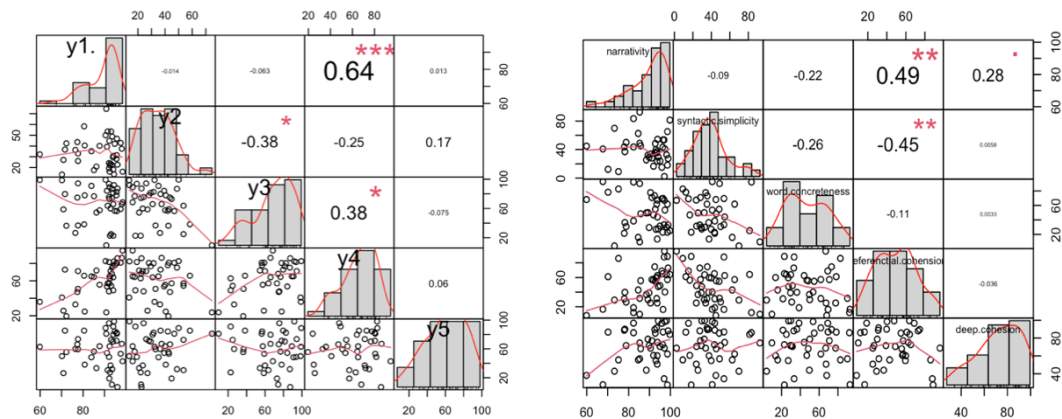

Figure 3. Correlations of discourse components for ChatGPT (left) and CIE learners (right)

## 4. Discussion

The present study using Coh-Metrix compared the writing performances by ChatGPT and CIE learners in narrative writing across five discourse components (i.e., narrativity, syntactic simplicity, word concreteness, referential cohesion, and deep cohesion) so as to reveal the chatbot's advantage and disadvantage in writing. Results indicated that ChatGPT performed better than CIE learners in terms of narrativity, word concreteness, and referential cohesion, but worse in syntactic simplicity and deep cohesion. After the revision upon the command given by human being, ChatGPT still performed worse than CIE learners in deep cohesion. All these answer the first and second question in the introduction. Additionally, narrativity was positively correlated with referential cohesion for both ChatGPT and CIE learners' groups. However, some differences were observed in the correlations between word concreteness and syntactic simplicity with referential cohesion for ChatGPT and CIE learners' groups. Specifically, word concreteness was negatively correlated with syntactic simplicity but positively correlated with referential cohesion for the ChatGPT group, while syntactic simplicity was negatively correlated with referential cohesion for the CIE learners' group. This result answers the third question posed in the introduction.

What follows elaborates on the potential aspects that may have motivated the results from the five discourse components and their correlations for narrative writing.

### 4.1 Narrativity

The statistical analysis indicated that ChatGPT outperformed CIE learners in terms of narrativity in either the initial or the revised version of writing tasks. First of all, this result may arise from the overuse of the first-person pronoun "*I*" in CIE learners' writing, as noted by Petchprasert (2021). CIE learners consistently used "*I*" throughout most of the sentences, suggesting a heavy emphasis on personal experiences related to the topic. However, using the first-person pronoun excessively can be considered as informal and inappropriate, as it may undermine the objectivity and credibility of the work (Li, 2014).

Another reason for ChatGPT's superiority in narrativity is probably due to the use of familiar and easy-to-understand language. Narrativity reflects the extent to which a story is familiar to readers and can be easily comprehended (Dela Rosa & Genuino, 2018; Kremzer, 2021). In CIE learners' writing, unexpected events such as near-death experiences were described using long and opaque words that could be unfamiliar and difficult to understand for many readers. In contrast, ChatGPT tended to employ more plain and straightforward words to describe personal experiences that occur frequently in people's daily life, such as public speaking and game playing. As a result, CIE learners received a lower score in narrativity compared to ChatGPT.

**4.2 Syntactic simplicity**

Although CIE learners outdid ChatGPT in terms of syntactic simplicity in the initial version, the revised version by ChatGPT showed similar performance relative to CIE learners in the revised version. However, it is important to note that syntactic simplicity is not always a reliable indicator of good or bad writing. Broadhead et al. (1982) proposed two measures that he believed indicate good writing: the length of base clauses (the shorter, the better) and the percentage of words in free modifiers (the higher percentages, the better, especially when in "final position" after a base clause). In our study, the sentences generated by the first version of ChatGPT contained an average of 18.15 words (SD = 6.17), ranging from 15.67 to 20.93 words, while CIE learners produced sentences containing an average of 15.43 words (SD = 3.14), ranging from 8 to 21 words. Apparently, in view of Broadhead et al. (1982), it looks as if CIE learners performed a bit better than or at least as equally good as ChatGPT in syntax.

Moreover, almost all the sentences in the first version by ChatGPT employed coordinating conjunctions "and" to connect two small sentences, and there were no free modifiers in the sentence, e.g., "*I met other solo travelers from around the world, and we shared our stories and experiences over meals and drinks*". In contrast, CIE learners tended to combine both long and short sentences in their writings and place modifiers in the final position of the sentence, e.g., "*Mom interrupted me. She boiled with anger and said: I am so regretful that you do not value your life.*" This phenomenon may be due to the fact that English majors are often encouraged by writing teachers to combine both long and short sentences to create better writing (Saddler, 2007). Short sentences increase readability, while long sentences can be used to emphasize key information and convey complex ideas. This suggests that the writings produced by the first version of ChatGPT had lower syntactic simplicity due to longer sentences and fewer words in free modifiers.

**4.3 Word concreteness**

The statistical data revealed that the two versions by ChatGPT outperformed CIE learners in word concreteness. This appears to remind the audience of the difference in abstraction of things by the chatbot and human writers: ChatGPT did not think deep of event-associated matter and hence was only good at logic-based (e.g., in terms of spatial order or chronological order) narration of storytelling or human experiences, while the CIE learners were not only good at narrating specific events but also did better in describing abstract ideas like metaphor/metonymy-based concepts.

It is widely known that the excessive use of abstract words leads to low word concreteness, and CIE learners tended to use more abstract words and ideas in their narrative writing. For example, one of the participants described how to prepare for the College Entrance Examination in China, using many abstract concepts (e.g., confidence, motivation, keeping silent) to express their mental activities before the examination, which was based on the abstraction of their background knowledge (Petchprasert, 2021). By contrast, when ChatGPT described the experience of finding a job, they utilized many action words (e.g., perfecting the resume and cover letter, researching the company, and preparing for potential interviews) to describe the actions taken during the job search process.

Furthermore, the higher word concreteness in ChatGPT's writing may be attributed to the diverse range of texts utilized in its training. ChatGPT was trained on various types of texts obtained from the internet, such as books, articles, and websites, covering a wide range of topics, including news, Wikipedia, and works of fiction (Yang et al., 2023). These resources used active words to describe events objectively and avoided the use of abstract words to express personal opinions or ideas. As a result, ChatGPT generated more objective and concrete words in its writing.

**4.4 Referential cohesion**

Our data revealed that ChatGPT, no matter whether in its first or in the revised version, had higher referential cohesion than CIE learners. This suggests that CIE learners' writing contained fewer explicit links between ideas, making it more difficult to read. There are likely three reasons for this phenomenon as below.

Firstly, as claimed by Kosinski (2023), ChatGPT exhibited a high level of theory of mind (ToM), which is related to referential cohesion. It is because that both ToM and referential cohesion require an understanding of the intentions and beliefs of others. To establish referential cohesion, speakers or writers need to make assumptions about the knowledge and beliefs of their listeners or readers, which requires a well-developed ToM and then use language to connect ideas in a way that makes sense to them. Yet the CIE learners do not have the equally good ToM in writing English as a second language as they have shown in using their native Chinese. It is no wonder that ChatGPT performs well in referential cohesion.

Secondly, CIE learners tended to repeat pronouns that had already been mentioned, as noted by Petchprasert (2021). For example, in one passage written by a major, the pronoun "we" was used three times in a single paragraph. This repetition created redundancy and was taken as a flaw since the writer did not clarify which noun "we" refers to. In contrast, ChatGPT never used pronouns such as "we/they/I" more than twice, which indicates a good awareness of diction diversity.

Thirdly, some majors struggled to use distinctive links to connect their ideas. For example, one passage contained the following sentence: "*The memory was formed when I was eight, just around the corner I turned nine, from a story that my younger sister and I had been lost in an overcast, dull afternoon.*" While the CIE writers may have been trying to create a certain atmosphere, the relationship between the memory and the story is unclear because there are no explicit associations between them required for logical connections. This makes it more difficult for readers to understand the passage.

**4.5 Deep cohesion**

Our data also revealed that CIE learners had a higher percentile in deep cohesion than ChatGPT in both the first and revised versions. Deep cohesion pertains to the quality and quantity of connecting words in texts, which are essential for constructing plot, pacing, and character development in narrative writing. Although inadequate cohesion can force readers to establish connections (Crossley and McNamara, 2014), the narrative writing in our study needs more connectives to construct plot, pacing, and character development. According to the statistics, ChatGPT's revised version lacked casual connectives (e.g., therefore, although) to make a sequence of sentences (the average of 20.29 connectives) but tended to use "and" (repeating more than 15 times) in 40 texts. To illustrate, ChatGPT misused coordinating conjunction 'and' rather than apply conjunctive adverb 'so/therefore' to connect the two casual sentences, e.g., "*I knew that the reward at the end of the hike would be worth the effort, and I was determined to reach the summit*". In addition, a majority of the texts from ChatGPT displayed a limited use of conjunctive adverbs (such as "still" and "even though") to create cohesive connections between sentences. For example, ChatGPT used "and" to connect "*I struggled to keep up with my friends*" and "*I felt a growing sense of frustration and exhaustion*". However, it is better to use "*but I still I felt a growing sense of ...*" to continue the previous sentence.

In contrast, CIE learners had more casual connectives (the average of 26.27 connectives) in their writings and they were good at using connectives to clearly show the logic relationships between sentences. For instance, in one participant's writing, a variety of conjunctive adverbs (e.g., as a result, therefore, thus) were adopted through the whole writing.

The difference in the usage of connectives (i.e., deep cohesion) by ChatGPT and CIE learners may be due to the limitations of text generation models (Zhao et al., 2022). Current text generation models perform poorly in terms of coherence, as the generated text often lacks logical flow and coherence, making it difficult for readers to understand or follow the intended message. Despite recent advancements in NLP, the lack of coherence in generated text remains a significant challenge for text generating models. This is because these models rely on statistical patterns and do not have a full understanding of the context, the meaning of the text in particular. As a consequence, improving coherence in text generation is an ongoing area in NLP research including ChatGPT.

**4.6 Correlations of discourse components in ChatGPT and CIE learners**

Our data analysis showed that narrativity was linked to referential cohesion in both the ChatGPT and CIE learners' groups, which supports Allen et al. 's (2019) research. Narratives usually necessitate a high degree of referential cohesion to effectively convey the story (Cross et al., 2016). To tell a coherent story, the different events and characters must be connected in a clear and easy-to-follow fashion for the reader. This often entails using cohesive devices to create and maintain links between different sections of the text. To the opposite, texts with low narrativity, such as academic or technical writing, may place less emphasis on referential cohesion because the focus is on conveying information rather than telling a story. In such texts, the use of cohesive devices may be more restricted because the goal is to present information in a clear and concise manner rather than establish a narrative structure.

Furthermore, the results showed that for the ChatGPT group, word concreteness had a negative correlation with syntactic simplicity, as concrete words generally have more complicated syntactic structures (Bulté & Housen, 2014; Polio & Shea, 2014). Concrete words often represent specific, tangible objects or concepts that may require more complex grammatical structures to accurately convey their meaning whereas abstract words represent less tangible concepts that may be easier to convey using simpler grammatical structures. However, word concreteness was positively correlated with referential cohesion. This is because concrete words are typically more precise and specific. This specificity can help to establish and maintain links between different parts of a text, as concrete words are easier to echo back and become visualizable. This, in turn, can aid in creating a text that is more cohesive and coherent by establishing clear connections between the different parts of writing.

Beyond our expectation, there came up syntactic simplicity having a negative correlation with referential cohesion in the CIE learners. This may be because more complex sentence structures often allow for a greater range of cohesive devices, which can aid in creating and maintaining connections between different parts of the text (Petchprasert, 2021). In other words, complex sentence structures may be necessary to convey complex ideas or relationships between concepts, which may require the use of more sophisticated cohesive devices. However, it is worth noting that the relationship between syntactic simplicity and referential cohesion is not always negative (Graesser et al., 2011). Depending on the context and purpose of writing, simpler sentence structures may be more appropriate and effective in establishing connections between different parts of the text.

**Conclusion**

In conclusion, this study compared the capacities of ChatGPT and CIE learners in generating narrative texts from the perspective of five discourse components. While ChatGPT initially outperformed CIE learners in some areas, it lagged far behind human writers in many other aspects, and further revision commands only improved certain aspects (e.g., syntactic simplicity) of its writing quality. Additionally, the findings suggest that ChatGPT with advanced ToM capabilities could potentially enhance their writing performance, especially in terms of referential cohesion.

This study suggests that AI tools like ChatGPT are really powerful in text writing but also lags far behind human writers particularly in the use of advanced inference like deep cohesion and underscores the need for continued research and development to optimize the performance of AI-powered writing tools.

Table 3. Five components of the first and revised version of writing from ChatGPT and CIE learners

| Variable | First version of ChatGPT | | Revised version of ChatGPT | | CIE learners | | F(2, 117) | p | η²p |
|---|---|---|---|---|---|---|---|---|---|
| | M | SD | M | SD | M | SD | | | |
| Narrativity | 92.294 | 4.983 | 87.60 | 9.448 | 88.611 | 9.640 | 3.539 | .032 | .057 |
| Syntactic simplicity | 30.974 | 12.270 | 33.899 | 13.751 | 40.252 | 19.982 | 3.654 | .029 | .059 |
| Word concreteness | 65.904 | 20.749 | 66.808 | 23.997 | 48.070 | 24.435 | 8.357 | <.001 | .125 |
| Referential cohesion | 73.155 | 15.150 | 62.144 | 19.483 | 47.431 | 23.271 | 17.372 | <.001 | .229 |
| Deep cohesion | 54.610 | 22.030 | 58.849 | 23.775 | 70.918 | 20.276 | 5.878 | .004 | .091 |